# A FACE RECOGNITION SCHEME USING WAVELET-BASED DOMINANT FEATURES

Hafiz Imtiaz and Shaikh Anowarul Fattah

Department of Electrical & Electronic Engineering, Bangladesh University of Engineering and Technology, Dhaka-1000, Bangladesh
hafizimtiaz@eee.buet.ac.bd, sfattah@princeton.edu

## ABSTRACT

*In this paper, a multi-resolution feature extraction algorithm for face recognition is proposed based on two-dimensional discrete wavelet transform (2D-DWT), which efficiently exploits the local spatial variations in a face image. For the purpose of feature extraction, instead of considering the entire face image, an entropy-based local band selection criterion is developed, which selects high-informative horizontal segments from the face image. In order to capture the local spatial variations within these high-informative horizontal bands precisely, the horizontal band is segmented into several small spatial modules. Dominant wavelet coefficients corresponding to each local region residing inside those horizontal bands are selected as features. In the selection of the dominant coefficients, a threshold criterion is proposed, which not only drastically reduces the feature dimension but also provides high within-class compactness and high between-class separability. A principal component analysis is performed to further reduce the dimensionality of the feature space. Extensive experimentation is carried out upon standard face databases and a very high degree of recognition accuracy is achieved by the proposed method in comparison to those obtained by some of the existing methods.*

## 1. INTRODUCTION

Automatic face recognition has widespread applications in security, authentication, surveillance, and criminal identification. Conventional ID card and password based identification methods, although very popular, are no more reliable as before because of the use of several advanced techniques of forgery and password-hacking. As an alternative, biometric, which is defined as an intrinsic physical or behavioural trait of human beings, is being used for identity access management [5]. The main advantage of biometric features is that these are not prone to theft and loss, and do not rely on the memory of their users. Among physiological biometrics, face is getting more popularity because of its non-intrusiveness and high degree of security. Moreover, unlike iris or finger-print recognition, face recognition do not require high precision equipment and user agreement, when doing image acquisition, which make face recognition even more popular for video surveillance.

Nevertheless, face recognition is a complicated visual task even for humans. The primary difficulty in face recognition arises from the fact that different images of a particular person may vary largely, while images of different persons may not necessarily vary significantly. Moreover, some aspects of the image, such as variations in illumination, pose, position, scale, environment, accessories, and age differences, make the recognition task more complicated. However, despite many relatively successful attempts to implement face recognition systems, a single approach, which is capable of addressing the hurdles, is yet to be developed.

The objective of this paper is to develop a wavelet-based face recognition scheme, which, instead of the entire face image, considers only some high-informative local zones of the image for dominant feature extraction. An entropy based horizontal band selection criterion is developed to exploit the high-informative areas of a face image. In order to precisely capture the local spatial variation within a high-informative horizontal band, such high-informative bands are further divided into some smaller spatial modules. We propose to extract dominant wavelet coefficients corresponding to some smaller segments residing within the band based on a threshold criterion. In comparison to the discrete Fourier transform, the DWT is used as it possesses a better space-frequency localization. It is shown that the discriminating capabilities of the proposed features are enhanced because of modularization of the face images. In view of further reducing the computational complexity, principal component analysis is performed on the proposed feature space. Finally, the face recognition task is carried out using a distance based classifier.

## 2. RELATED WORKS

Face recognition methods are based on extracting unique features from face images. In this regard, face recognition approaches can be classified into two main categories: holistic and texture-based [13]-[15]. Holistic or global approaches to face recognition involve encoding the entire facial image in a high-dimensional space [13], [18]. It is assumed that all faces are constrained to particular positions, orientations, and scales. However, texture-based approaches rely on the detection of individual facial characteristics and their geometric relationships prior to performing face recognition [11], [15], [17]. Apart from these approaches, face recognition can also be performed by using different local regions of face images [2]-[4]. It is well-known that, although face images are affected due to variations, such as non-uniform illumination, expressions and partial occlusions, facial variations are confined mostly to local regions. It is expected that capturing these localized variations of images would result in a better recognition accuracy [4], [16]. In this regard, wavelet analysis is also employed that possesses good characteristics of spatial-frequency localization to detect facial geometric structure [8]. Because of the property of shift-invariance, it is well known that wavelet based approach is one of the most robust feature extraction schemes, even under variable illumination [12]. Hence, it is motivating to utilize local variations of face geometry using wavelet transform for feature extraction and thereby develop a face recognition scheme incorporating the advantageous properties of both holistic- and texture-based approaches.

## 3. BRIEF DESCRIPTION OF THE PROPOSED SCHEME

A typical face recognition system consists of some major steps, namely, input face image collection, pre-processing, feature extraction, classification and template storage or database, as illustrated in Fig. 1. The input image can be collected generally from a video camera or still camera or surveillance camera. In the process of capturing images, distortions including rotation, scaling, shift and translation may be present in the face images, which make it difficult to locate at the correct position. Pre-processing removes any un-wanted objects (such as, background) from the collected image. It may also segment the face image for feature extraction. For the purpose of classification, an image database is needed to be prepared consisting template face poses of different persons. The recognition task is based on comparing a test face image with template data. It is obvious that considering images themselves would require extensive computations for the purpose of comparison. Thus, instead of utilizing the raw face images, some characteristic features are extracted for preparing the template. It is to be noted that the recognition accuracy strongly depends upon the quality of the extracted features. Therefore, the main focus of this research is to develop an efficient feature extraction algorithm.

The proposed feature extraction algorithm is based on extracting spatial variations precisely from high informative local zones of the face image instead of utilizing the entire image. In view of this, an entropy based selection criterion is developed to select high informative facial zones. A modularization technique is employed then to segment the high informative zones into several smaller segments. It should be noted that variation of illumination of different face images of the same person may affect their similarity. Therefore, prior to feature extraction, an illumination adjustment step is included in the proposed algorithm. After feature extraction, a classifier compares features extracted from face images of different persons and a database is used to store registered templates and also for verification purpose.

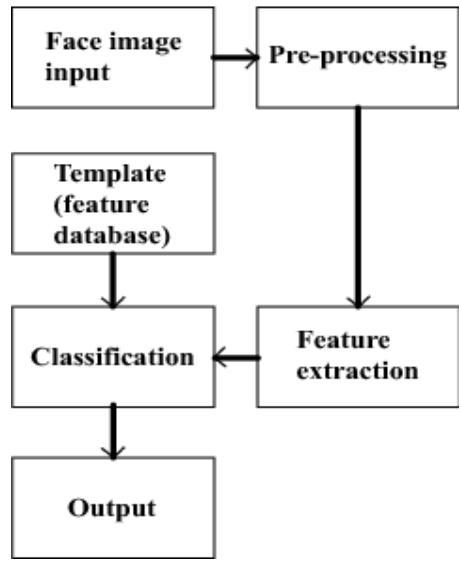

Figure  1: Block diagram of the proposed method

# 4. PROPOSED METHOD

For any type of biometric recognition, the most important task is to extract distinguishing features from the training biometric traits, which directly dictates the recognition accuracy. In comparison to person recognition based on different biometric features, face image based recognition is very challenging even for a human being, as face images of different persons may seem similar whereas face images of a single person may seem different, under different conditions. Thus, obtaining a significant feature space with respect to the spatial variation in a human face image is very crucial. In what follows, we are going to demonstrate the proposed feature extraction algorithm for face recognition, where spatial domain local variation is extracted using wavelet domain transform.

## 4.1  Entropy Based Horizontal Band Selection

The information content of different regions of a human face image vary widely [1]. It can be shown that, if an image of a face were divided into certain segments, not all the segments would contain the same amount of information. It is expected that a close neighborhood of eyes, nose and lips contains more information than that possessed by the other regions of a human face image. It is obvious that a region with high information content would be the region of interest for the purpose of feature extraction. However, identification of these regions is not a trivial task. Estimating the amount of information from a given image can be used to identify those significant zones. In this paper, in order to determine the information content in a given area of a face image, an entropy based measure of intensity variation is defined as [9]

$$H = -\sum_{k=1}^{m} p_k \ log_2 p_k, \qquad\qquad (1)$$

where the probabilities $\{p_k\}_1^m$ are obtained based on the intensity distribution of the pixels of a segment of an image. It is to be mentioned that the information in a face image exhibits variations more prominently in the vertical direction than that in the horizontal direction [3]. Thus, the face image is proposed to be divided into several horizontal bands and the entropy of each band is to be computed. It has been observed from our experiments that variation in entropy is closely related to variation in the face geometry. Fig. 2(b) shows the entropy values obtained in different horizontal bands of a person for several sample face poses. One of the poses of the person is shown in Fig. 2(a). As expected, it is observed from the figure that the neighborhood of eyes, nose and lips contains more information than that possessed by the other regions. Moreover, it is found that the locus of entropies obtained from different horizontal bands can trace the spatial structure of a face image. Hence, for feature extraction in the proposed method, spatial horizontal bands of face images are chosen corresponding to their entropy content.

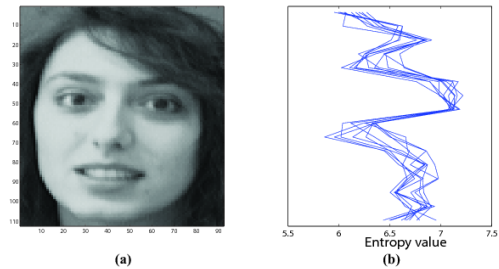

Figure 2: (a) Sample face image of a person and (b) entropy values in different horizontal bands of several face poses

## 4.2 DWT-based Feature Extraction and Illumination Adjustment

For biometric recognition, feature extraction can be carried out using mainly two approaches, namely, the spatial domain approach and the frequency domain approach [14]. The spatial domain approach utilizes the spatial data directly from the face image or employs some statistical measure of the spatial data. On the other hand, frequency domain approaches employ some kind of transform over the face images for feature extraction. In case of frequency domain feature extraction, pixel-by-pixel comparison between face images in the spatial domain is not necessary. Phenomena, such as rotation, scale and illumination, are more severe in the spatial domain than in frequency domain. Hence, in what follows, we intend to develop a feature extraction algorithm based on multi-resolution transformation.

Since it is shown in the previous section that certain zones of a face image consist considerably higher information in comparison to other zones, unlike conventional approaches, our objective is to extract features from the spatial data residing only in the high-informative facial bands. Obviously, such a method of feature extraction reduces the feature dimension, which results in significant computational savings. For feature extraction, we have employed 2D-DWT, which, in comparison to the Fourier transform, possesses a better space-frequency localization. This property of the DWT is helpful for analyzing images, where the information is localized in space. The wavelet transform is analogous to the Fourier transform with the exception that it uses scaled and shifted versions of wavelets and the decomposition of a signal involves sum of these wavelets. The DWT kernels exhibit properties of horizontal, vertical and diagonal directionality. The continuous wavelet transform (CWT) of a signal $s(t)$ using a wavelet $\psi(t)$ is mathematically defined as

$$C(a,b) = \frac{1}{\sqrt{a}} \int s(t) \psi(\frac{t-b}{a}) dt, \qquad (2)$$

where $a$ is the scale and $b$ is the shift. The DWT coefficients are obtained by restricting the scale $(a)$ to powers of $2$ and the position $(b)$ to integer multiples of the scales, and are given by

$$c_{j,k} = 2^{j/2} \int_{-\infty}^{\infty} s(t) \psi(2^j t - k) dt, \qquad (3)$$

where $j$ and $k$ are integers and $\psi_{j,k}$ are orthogonal baby wavelets defined as

$$\psi_{j,k} = 2^{j/2} \psi(2^j t - k) \qquad (4)$$

The approximate wavelet coefficients are the high-scale low-frequency components of the signal, whereas the detail wavelet coefficients are the low-scale high-frequency components. The 2D-DWT of a two-dimensional data is obtained by computing the one-dimensional DWT, first along the rows and then along the columns of the data. Thus, for a 2D data, the detail wavelet coefficients can be classified as vertical, horizontal and diagonal detail.

It is intuitive that images of a particular person captured under different lighting conditions may vary significantly, which can affect the face recognition accuracy. In order to overcome the effect of lighting variation in the proposed method, illumination adjustment is performed prior to feature extraction. Given two images of a single person having different intensity distributions due to variation in illumination conditions, our objective is to provide with similar feature vectors for these two images irrespective of the different illumination condition. Since in the proposed method, feature extraction is performed in the DWT domain, it is of our interest to analyze the effect of variation in illumination on the DWT-based feature extraction.

In Fig. 3, two face images of the same person are shown, where the second image (shown in Fig. 3(b) is made brighter than the first one by changing the average illumination level. 2D-DWT is performed upon each image, first without any illumination adjustment and then after performing illumination adjustment. Considering all the 2D-DWT approximate coefficients to form the feature vectors for these two images, a measure of similarity can be obtained by using correlation. In Figs. 4 and 5, the cross-correlation values of the 2D-DWT approximate coefficients obtained by using the two images without and with illumination adjustment are shown, respectively. It is evident from these two figures that the latter case exhibits more similarity between the DWT approximate coefficients indicating that the features belong to the same person. The similarity measure in terms of Euclidean distances between the 2D-DWT approximate coefficients of the two images for the aforementioned two cases are also calculated. It is found that there exists a huge separation in terms of Euclidean distance when no illumination adjustment is performed, whereas the distance completely diminishes when illumination adjustment is performed, as expected, which clearly indicates that a better similarity between extracted feature vectors.

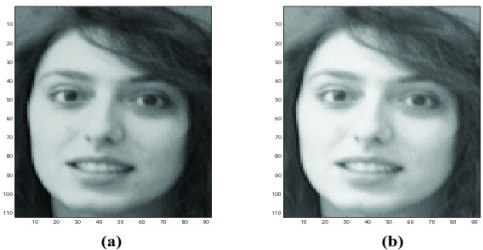

(a)            (b)

Figure 3: Two face images of the same person under different illumination

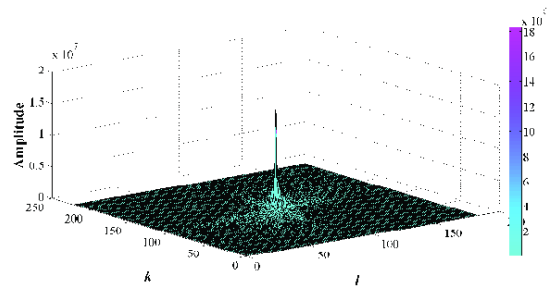

Figure 4: Correlation of the 2D-DWT approximate coefficients of the sample images: no illumination adjustment

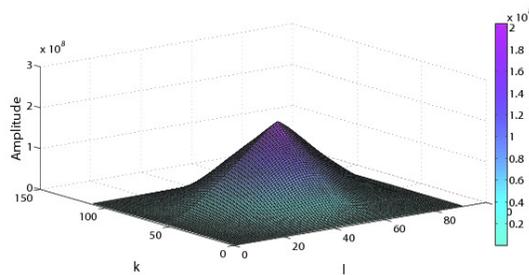

Figure 5: Correlation of the 2D-DWT approximate coefficients of the sample: illumination adjusted

### 4.3 Proposed Wavelet Domain Dominant Feature

Instead of considering the DWT coefficients of the entire image, the coefficients obtained form each modules of the high-informative horizontal band of a face image are considered to form the feature vector of that image. However, if all of these coefficients were used, it would definitely result in a feature vector with a very large dimension. In view of reducing the feature dimension, we propose to utilize the dominant wavelet coefficients as desired features. Wavelet coefficients, which are greater than a certain threshold value, are treated as the dominant coefficients. It is intuitive that within a high-informative horizontal band of a face image, the image intensity distribution may drastically change at different localities. In order to select the dominant wavelet coefficients, if the thresholding operation were to be performed over the wavelet coefficients of the entire band, it would be difficult to obtain a global threshold value that is suitable for every local zone. Use of a global threshold in a particular horizontal band of a face image may offer features with very low between-class separation. In order to obtain high within-class compactness as well as high between-class separability, we have considered wavelet coefficients corresponding to some smaller spatial modules residing within a horizontal band, which are capable of extracting variation in image geometry locally. In this case, for each module, a different threshold value may have to be chosen depending on the coefficient values of that segment. The wavelet coefficients (both approximate and horizontal detail) of a segment of the high-informative horizontal band are sorted in descending order and the top $\theta\%$ of the coefficients are considered as dominant wavelet coefficients and selected as features for the particular segment of the horizontal band. This operation is repeated for all the modules of a high-informative horizontal band.

Next, in order to demonstrate the advantage of extracting dominant wavelet coefficients corresponding to some smaller modules residing in a horizontal band, we conduct an experiment considering two different cases: (*i*) when the entire horizontal band is used as a whole and (*ii*) when all the modules of that horizontal band are used separately for feature extraction. For these two cases, centroids of the dominant approximate wavelet coefficients obtained from several poses of two different persons (appeared in Fig. 6) are computed and shown in Figs. 7 and 8, respectively. It is observed from Fig. 7 that the feature-centroids of the two persons at different poses are not well-separated and even for some poses they overlap with each other, which clearly indicates poor between-class separability. In Fig. 8, it is observed that, irrespective of the poses, the feature-centroids of the two persons maintain a significant separation indicating a high between-class separability, which strongly supports the proposed local feature selection algorithm.

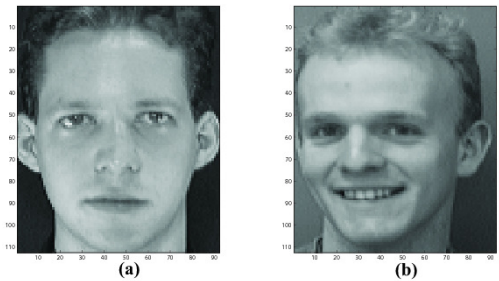

Figure 6: Sample face images of two persons

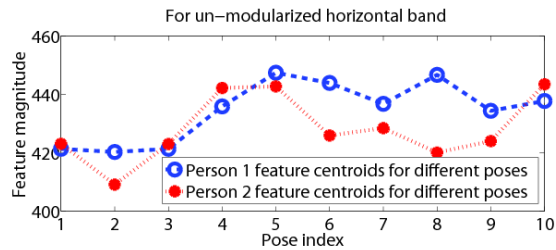

Figure 7: Feature centroids of different poses for un-modularized horizontal band

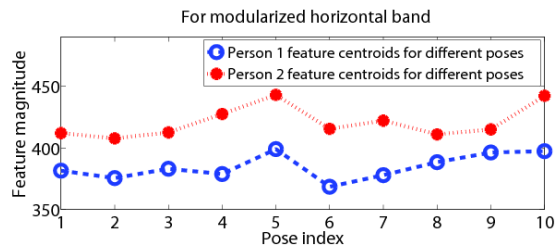

Figure 8: Feature centroids of different poses for modularized horizontal band

We have also considered dominant feature values obtained for various poses of those two persons in order to demonstrate the within class compactness of the features. The feature values, along with their centroids, obtained for the two different cases, i.e., extracting the features from the horizontal band without and with modularization, are shown in Figs. 9 and 10, respectively. It is observed from Fig. 9 that the feature values of several poses of the two different persons are significantly scattered around the respective centroids resulting in a poor within-class compactness. On the other hand, it is evident from Fig. 10 that the centroids of the dominant features of the two different persons are well-separated with a low degree of scattering among the features around their corresponding centroids. Thus, the proposed dominant features extracted locally within a band offer not only a high degree of between-class separability but also a satisfactory within-class compactness.

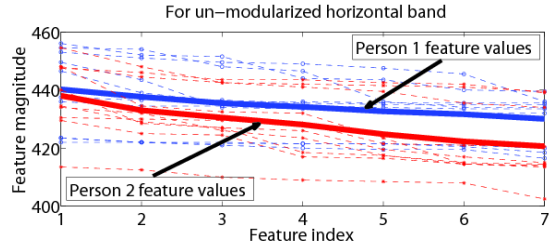

Figure 9: Feature values for un-modularized horizontal band

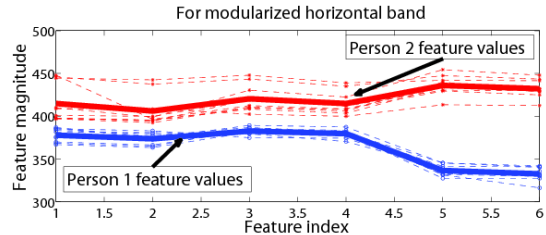

Figure 10: Feature values for modularized horizontal band

## 4.4 Reduction of the Feature Dimension

Principal component analysis (PCA) is an efficient orthogonal linear transform to reduce the feature dimension [7]. In the proposed method, considering the dominant 2D-DWT coefficients as features results in a feature space with large dimension. Thus, implementation of PCA on the derived feature space could efficiently reduce the feature dimension without loosing much information. Hence, PCA is employed to reduce the dimension of the proposed feature space.

## 4.5 Distance Based Face Recognition

In the proposed method, for the purpose of recognition using the extracted dominant features, a distance-based similarity measure is utilized. The recognition task is carried out based on the distances of the feature vectors of the training face images from the feature vector of the test image. Given the $m$-dimensional feature vector for the $k$-th pose of the $j$-th person be $\{\gamma_{jk}(1), \gamma_{jk}(2), \ldots, \gamma_{jk}(m)\}$ and a test face image $f$ with a feature vector $\{v_f(1), v_f(2), \ldots, v_f(m)\}$, a similarity measure between the test image $f$ of the unknown person and the sample images of the $j$-th person, namely *average sum-squares distance, $\Delta$*, is defined as

$$\Delta_j^f = \frac{1}{q} \sum_{k=1}^{q} \sum_{i=1}^{m} |\gamma_{jk}(t) - v_f(t)|^2, \qquad (5)$$

where a particular class represents a person with $q$ number of poses. Therefore, according to (5), given the test face image $f$, the unknown person is classified as the person $j$ among the $p$ number of classes when

$$\Delta_j^f \leq \Delta_g^f, \quad \forall j \neq y \quad and \quad \forall y \varepsilon \{1, 2, \dots, p\} \quad (6)$$

## 5. EXPERIMENTAL RESULTS

Extensive simulations are carried out in order to demonstrate the performance of the proposed feature extraction algorithm for face recognition. In this regard, different well-known face databases have been considered, which consist a range of different face images varying in facial expressions, lighting effects and presence/absense of accessories. The performance of the proposed method in terms of recognition accuracy is obtained and compared with that of some recent methods [10, 6].

### 5.1 Face Databases

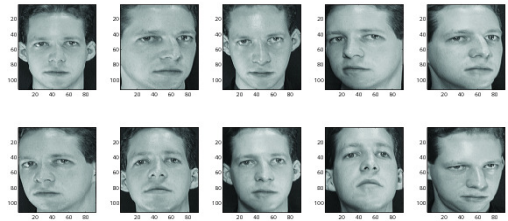

Figure 11: Sample poses of a person from the ORL database

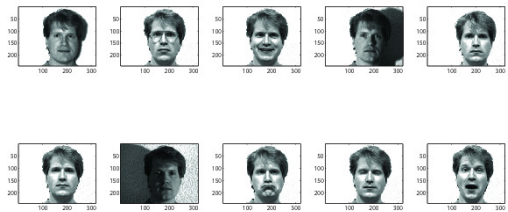

Figure 12: Sample poses of a person from the Yale database

In this section, the performance of the proposed face recognition scheme has been presented for two standard face databases, namely, the ORL database (available at http://www.cl.cam.ac.uk/Research/DTG/attarchive/pub/data/) and the Yale database (available at http://cvc.yale.edu/projects/yalefaces/yalefaces.html). In Figs. 11 and 12, sample face images of different poses of two different persons taken from the ORL and the Yale databases, respectively, are shown. The ORL database contains a total of 400 images of 40 persons, each person having 10 different poses. Little variation of illumination, slightly different facial expressions and details are present in the face images. The Yale database, on the other hand, consists a total of 165 images of 15 persons, each person having 11 different poses. The poses exhibit large variations in illumination (such as central lighting, left lighting and right lighting, dark condition), facial expressions (such as wink, sad, happy, surprised, sleepy and normal) and other accessories (such as with glasses and without glass).

## 5.2  Performance Comparison

In the proposed method, dominant features (approximate and horizontal detail 2D-DWT coefficients) obtained from all the modules of high-informative horizontal bands of a face image are used to form the feature vector of that image and feature dimension reduction is performed using PCA. The recognition task is carried out using a simple Euclidean distance based classifier as described in Section 3.5. The experiments were performed following the leave-one-out cross validation rule.

For simulation purposes, $N$ number of horizontal bands are selected based on the entropy measure described in Section 3.1 and divided further into small modules. Module height is the same as that of the horizontal band and module width is chosen based on the face image width. In our simulations, $N = 2$ for the ORL database and $N = 3$ for the Yale database are chosen and the module sizes are chosen as $16 \times 16$ pixels and $32 \times 32$ pixels, respectively. The dominant wavelet coefficients corresponding to all the local segments residing in the horizontal bands are then obtained using $\theta = 20$.

For the purpose of comparison, recognition accuracies obtained using the proposed method along with those obtained by the methods reported in [10] and [6] are listed in Table 1. Here, in case of the ORL database, the recognition accuracy for the method in [6] is denoted as not available (N/A). It is evident from the table that the recognition accuracy of the proposed method is comparatively higher than those obtained by the other methods for both the databases. It indicates the robustness of the proposed method against partial occlusions, expressions and nonlinear lighting variations.

Table  1: Comparison of recognition accuracies

| Method | Yale database | ORL database |
|---|---|---|
| Proposed method | 98.71% | 99.75% |
| Method [10] | 98.18% | 99.00% |
| Method [6] | 97.70% | N/A |

## 6. CONCLUSIONS

The proposed wavelet-based dominant feature extraction algorithm provides an excellent space-frequency localization, which is clearly reflected in the high within-class compactness and high between-class separability of the extracted features. Instead of using the whole face image for feature extraction at a time, first, certain high-informative horizontal bands within the image are selected using the proposed entropy based measure. Modularization of the horizontal bands is performed and the dominant wavelet coefficient features are extracted from within those local zones of the horizontal bands. It has been found that the proposed feature extraction scheme offers an advantage of precise capturing of local variations in the face images, which plays an important role in discriminating different faces. Moreover, it utilizes a very low dimensional feature space, which ensures lower computational burden. For the task of classification, an Euclidean distance based classifier has been employed and it is found that, because of the quality of the extracted features, such a simple classifier can provide a very satisfactory recognition performance and there is no need to employ any complicated classifier. From our extensive simulations on different standard face databases, it has been found that the proposed method provides high recognition accuracy even for images affected due to partial occlusions, expressions and nonlinear lighting variations.

## ACKNOWLEDGEMENTS

The authors would like to express their sincere gratitude towards the authorities of the Department of Electrical and Electronic Engineering and Bangladesh University of Engineering and Technology (BUET) for providing constant support throughout this research work.

# Author

**Hafiz Imtiaz** was born in Rajshahi, Bangladesh on January 16, 1986. He received his B.Sc. and M.Sc. degrees from Bangladesh University of Engineering and Technology (BUET), Dhaka, Bangladesh in March 2009 and July 2011, respectively, from the department of Electrical and Electronic Engineering (EEE). He is presently working as a lecturer in the department of EEE, BUET. He is also serving as the Treasurer of the IEEE Bangladesh Section.

**Shaikh Anowarul Fattah** S. A. Fattah received the B.Sc. and M.Sc. degrees from Bangladesh University of Engineering and Technology (BUET), Dhaka, Bangladesh, in 1999 and 2002, respectively, both in electrical and electronic engineering (EEE). He received the Ph.D. degree in electrical and computer engineering from Concordia University, Montreal, QC, Canada in 2008. He was a visiting Postdoctoral Fellow in the Department of Electrical Engineering at Princeton University, New Jersey, USA. From 2000 to 2003, he served as a Lecturer and later became an Assistant Professor in the Department of EEE, BUET. He is a recipient of the Dr. Rashid Gold Medal for the best academic performance in M.Sc. He was selected as one of the Great Grads of Concordia University in 2008-2009 and the winner of Concordia University's 2009 Distinguished Doctoral Dissertation Prize in Engineering and Natural Sciences. During his Ph.D. program, he has also received Concordia University Graduate Fellowship, Power Corporation of Canada Graduate Fellowship, a New Millennium Graduate Scholarship, Hydro Quebec Awards, International Tuition-fee Remission Award, and Doctoral Teaching Assistantship. He is the recipient of URSI Canadian Young Scientist Award 2007 and was the first prize winner in the SYTACOM Research Workshop in 2008. He was a recipient of NSERC Postdoctoral Fellowship in 2008. Dr. Fattah has published more than sixty international journal and conference papers. His research interests include the areas of system identification and modeling, speech, audio, and music signal processing, genomic signal processing, biomedical signal processing, biometric recognition for security, multimedia communication and control system. He is serving as technical committee members of different international conferences. He served as a member secretary of the EUProW 2009. He is the reviewer of different international and IEEE journals, such as IEEE TRANSACTIONS ON CIRCUITS AND SYSTEMS and IEEE TRANSACTIONS ON SIGNAL PROCESSING.